\documentclass[journal, 12pt, draftclsnofoot, onecolumn]{IEEEtran}
\usepackage{graphicx}
\usepackage{cite}
\usepackage{picinpar}
\usepackage{amsmath}
\usepackage{url}
\usepackage{flushend}
\usepackage[latin1]{inputenc}
\usepackage{colortbl}
\usepackage{soul}
\usepackage{multirow}
\usepackage{pifont}
\usepackage{color}
\usepackage{alltt}
\usepackage[hidelinks]{hyperref}
\usepackage{enumerate}
\usepackage{siunitx}
\usepackage{breakurl}
\usepackage{epstopdf}
\usepackage{pbox}

\usepackage{array}
\usepackage{threeparttable}
\usepackage{multirow}
\usepackage{algorithm}
\usepackage{algorithmic}
\usepackage{bm}
\usepackage{CJK}
\usepackage{indentfirst}
\usepackage{amsmath}
\usepackage{subfigure}
\usepackage{color}
\usepackage{booktabs}
\usepackage{amsmath,amssymb}
\usepackage{epstopdf}
\usepackage{textcomp}
\usepackage{diagbox}
\usepackage{setspace}

\usepackage{soul}
\soulregister\cite7 
\soulregister\citep7 
\soulregister\citet7
\soulregister\ref7 
\soulregister\pageref7
\begin{document}
\title{A Transfer Learning-based State of Charge Estimation for Lithium-Ion Battery at Varying Ambient Temperatures}

\author{Yan~Qin,
	Stefan~Adams,
	and Chau~Yuen, ~\IEEEmembership{Fellow,~IEEE}
\thanks{This work was supported by the A*STAR-NTU-SUTD Joint Research Grant on Artificial Intelligence Partnership under Grant RGANS1906 and in part supported by the National Natural Science Foundation of China under Grant 61903327. (Corresponding author: Yan Qin.)}
\thanks{Y. Qin and C. Yuen are with the Engineering Product Development Pillar, Singapore University of Technology and Design,  8 Somapah Road, 487372 Singapore. (e-mail: yan$\_$qin@sutd.edu.sg, yuenchau@sutd.edu.sg)}
\thanks{S. Adams is with the Department of Materials Science and Engineering, National University of Singapore, 117575 Singapore. (e-mail: mseasn@nus.edu.sg).}
}

\maketitle
\renewcommand{\baselinestretch}{2.0}
\begin{abstract}
Accurate and reliable state of charge (SoC) estimation becomes increasingly important to provide a stable and efficient environment for Lithium-ion batteries (LiBs) powered devices. Most data-driven SoC models are built for a fixed ambient temperature, which neglect the high sensitivity of LiBs to temperature and may cause severe prediction errors. Nevertheless, a systematic evaluation of the impact of temperature on SoC estimation and ways for a prompt adjustment of the estimation model to new temperatures using limited data have been hardly discussed. To solve these challenges, a novel SoC estimation method is proposed by exploiting temporal dynamics of measurements and transferring consistent estimation ability among different temperatures. First, temporal dynamics, which is presented by correlations between the past fluctuation and the future motion, is extracted using canonical variate analysis. Next, two models, including a reference SoC estimation model and an estimation ability monitoring model, are developed with temporal dynamics. The monitoring model provides a path to quantitatively evaluate the influences of temperature on SoC estimation ability. After that, once the inability of the reference SoC estimation model is detected, consistent temporal dynamics between temperatures are  selected for transfer learning. Finally, the efficacy of the proposed method is verified through a benchmark. Our proposed method not only reduces prediction errors at fixed temperatures (e.g., reduced by 24.35$\%$ at -20$^\circ$C, 49.82$\%$ at 25$^\circ$C) but also improves prediction accuracies at new temperatures.
\end{abstract}
\begin{IEEEkeywords}
Lithium-ion battery, SoC estimation, transfer learning, varying ambient temperature, long short-term memory network.
\end{IEEEkeywords}

\markboth{}%
{}

\definecolor{limegreen}{rgb}{0.2, 0.8, 0.2}
\definecolor{forestgreen}{rgb}{0.13, 0.55, 0.13}
\definecolor{greenhtml}{rgb}{0.0, 0.5, 0.0}

\section{Introduction}
\IEEEPARstart{W}{ith} merits of fast charging, high energy density, low cost, and long lifespan \cite{Ref1},\cite{Ref2}, Lithium-ion batteries (LiBs) provide a promising alternative to fossil fuels and have been applied to a wide range of applications, including electric vehicles (EV), personal mobility devices, portable instruments, smart-grid energy storage devices, etc. Accurate state of charge (SoC) estimation, which measures remaining capacity for LiBs [3], not only has a significant influence on the performance of its powered system, e.g., driving ranges of EVs, but also is important to assess the health status of LiBs and ensure a safe environment for LiBs powered systems.

Over the past decades, SoC estimation has attracted considerable attention both in industrial and academic community. In the early days, model-based methods \cite{Ref4},\cite{Ref5} have been widely used, which, however, require plentiful process knowledge that is time-consuming, expensive, and difficult to derive. Additionally, model-based methods are specific to details of the actual battery materials combinations and the fabrication process, resulting in poor generalization from one battery type to another.

With the rapid development in sensor and data acquisition technologies, a wealth of LiBs cycling data become available. Reflecting the inherent status using easily obtained data resources, data-driven SoC estimation methods have shown their superiorities in the era of big data and artificial intelligence. Taking current, voltage, and temperature as inputs, He et al. \cite{Ref6} proposed a method using artificial neural network to estimate SoC values. However, measurements were assumed to be time independent during model training, which might not hold true due to strong correlations between past and future samplings. To consider the time dependence of measurements, Li et al. \cite{Ref7} proposed an estimation method based on recurrent neural network (RNN), wherein hidden nodes were designed to memorize previous information by introducing feedback linkage. Although RNN outperforms traditional neural networks, it faces the vanishing descent problem due to an increasing burden of candidate states with the increase of time. Alternatively, long short-term memory (LSTM) network introduces a forget mechanism into the hidden nodes of RNN to overcome the vanishing descent problem by freeing memory from useless information. Chemali et al. \cite{Ref8} put forward a  LSTM-based method and achieved obvious improvements. However, a crucial implication of current methods [6]-[8] lies in that they are developed at a fixed ambient temperature.

Although researchers are fully motivated to develop more accurate SoC estimation models for LiBs, it is facing challenges due to various internal and external factors. Especially, LiBs are highly sensitive to ambient temperature [9], implying that the pre-trained model developed at a specific ambient temperature cannot work reliably at other ambient temperatures. However, it is common that the battery powered Internet of Thing devices or EVs have to operate over a wide range of ambient temperatures. For example, when an EV is driven in a long journey or rapidly (dis)charged, major ambient temperature changes may be experienced from day to night. The discharging efficiency will also drastically be reduced under extreme ambient temperatures, mainly low temperatures [10]. A SoC prediction system in the car that does not take this into account will show a poor prediction of SoC and driving range. Achievements dealing with varying temperatures with physical models have been reported [11], [12], however, they may suffer from the same challenges of model-based methods. To the best of the authors' knowledge, feasible solutions for data-driven SoC estimation have not been proposed yet.

Recently, transfer learning (TL) has been developed as a novel way to learn properties from source data and transfer them to the target data, which has achieved great success in image, audio, and text processing \cite{Ref13}. Inspired by this, it provides an opportunity for rapid modeling of SoC estimation from a comparatively small amount of data if estimation ability from known similar states (such as in this case other known temperatures) could be transferred to a new one. TL sequentially addresses three core questions: when to transfer, what to transfer, and how to transfer. Naturally, it raises three critical issues with respect to SoC estimation for LiBs taking temperature into account: 1) how to judge whether the previously developed model, i.e., the reference SoC estimation model, is still suitable or not; 2) once the estimation relationship has been changed, what kind of information in the reference model can be transferred; 3) how to achieve rapid SoC estimation modeling with less data at new temperatures.

In this paper, based on the in-depth analysis of LiBs data characteristics, the following recognition are gained: 1) temporal dynamics is similar within a small temperature region; 2) despite temporal dynamics may be dissimilar when temperature varies, consistent temporal dynamics, which remains the same at different temperatures, may exist due to the nature of the electrochemical process. Herein, the temporal dynamics refers to the correlations between past fluctuation and the future motion. Based on these recognitions, a TL-based SoC estimation method considering temporal dynamics is proposed. First, data processing is achieved through performing wavelet analysis on measurements to deal with the non-stationarity and multi-scale characteristics. Next, temporal correlations, which reflect the SoC estimation ability, are extracted using canonical variate analysis (CVA) as feature engineering. After that, canonical variates (CVs) are fed into LSTM network to develop the reference SoC estimation model, achieving higher prediction accuracy. The efficacy of estimation model in face of varying temperature is quantitatively evaluated by monitoring normal variation of CVs in a linear and explainable way. Furthermore, if the reference SoC estimation model becomes invalid, alarms will be issued promptly and a rapid model updating strategy is triggered by transferring consistent estimation ability. Estimation monitoring and rapid model updating minimize the required amount of new data and ensure a high adaptivity to temperature changes. The contributions of the proposed method are summarized below:
\begin{enumerate}[1)]
	\item A data-driven strategy is proposed for SoC estimation under varying temperatures;
	\item Estimation ability monitoring is designed to assess the efficacy of the reference model;
	\item  A rapid model updating strategy with less data is proposed to adapt to new temperatures via transferring consistent estimation ability.
\end{enumerate}

The rest of this article is structured as follows: Section II describes the discharge procedure and compares two machine learning (ML) methods used in this article. Next, details of the proposed methods are shown in Section III. In Section IV, efficacy of the proposed method is verified through a LiB benchmark. Finally, the work is concluded in the last section.
\section{Preliminary}
This section begins with the description of LiBs discharge procedure. After that, comparisons of two popular ML methods used in this article, i.e., CVA and LSTM, are given for better understanding.

\subsection{Discharging Procedure and Data Illustration}
Discharge procedure provides energy for LiBs powered devices. The available capacity is high at a full charge state, corresponding to a large value of SoC. Once discharging, SoC values will decrease until voltage signal reached a pre-set threshold. Fig. 1 depicts two widely adopted variables, i.e., current and voltage, which are obtained in an experimental discharge cycle at 25$^\circ$C under US06 dynamic testing of a Panasonic 18650PF cell \cite{Ref14}. Current signal varies between -20A to 10A, and voltage signal shows a descending trend. The true value of SoC is shown at the bottom of Fig. 1, which continuously decreases during the discharging.
\begin{figure}
	\centering
	\includegraphics[scale=0.55]{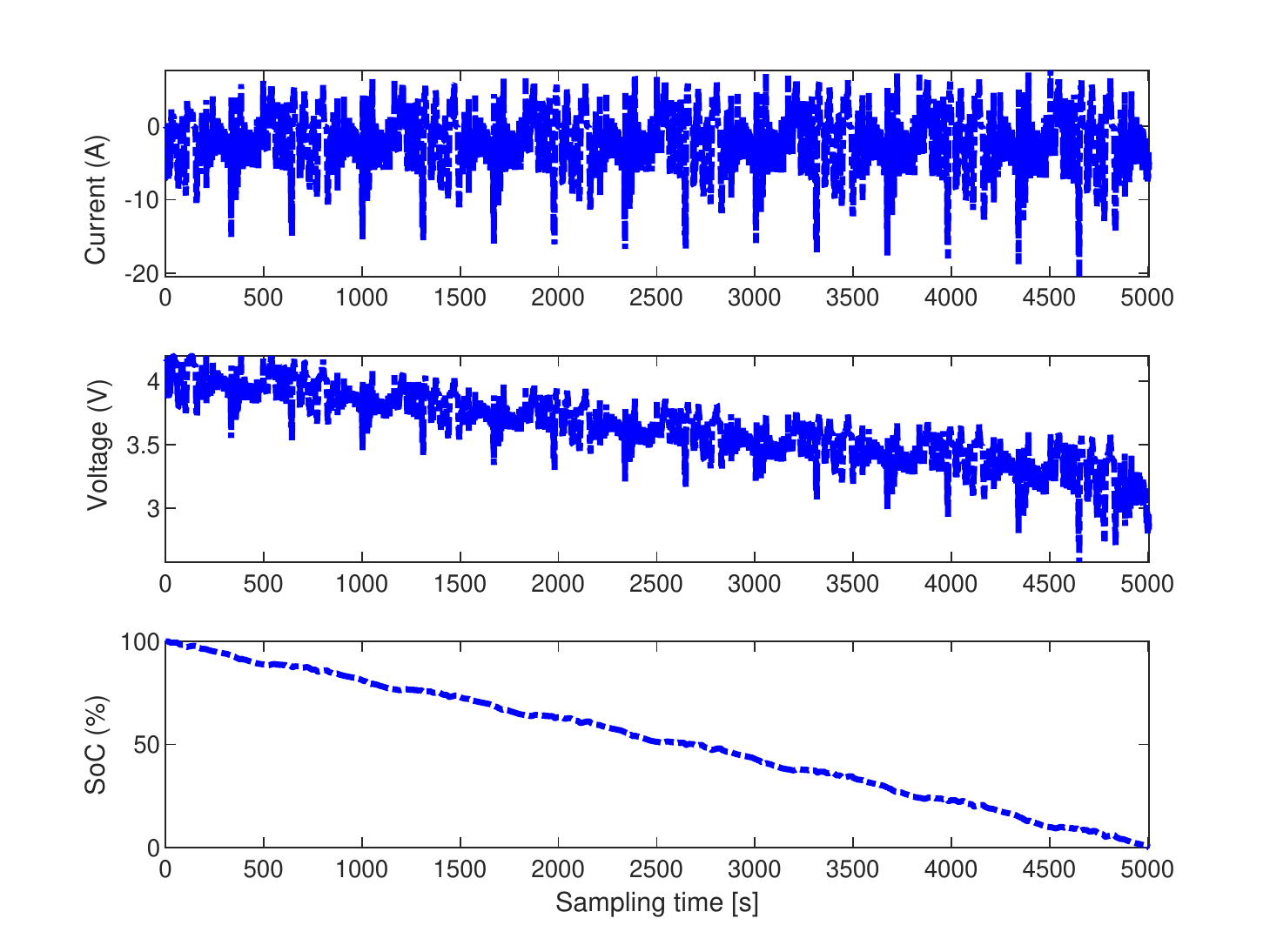}\\
	\caption{Measurements of a real discharge cycle with US06 testing at 25$^\circ$C of a Panasonic 18650PF cell \cite{Ref14}.}
	\vspace{-0.5cm}
	\label{MyFig1}
\end{figure}

\subsection{Comparisons between CVA and LSTM}
\begin{table*}[ht]
	\centering
	\tiny
	\renewcommand{\arraystretch}{1.2}
	\caption{Comprehensive comparisons between CVA and LSTM with respect to temporal dynamics.}
	\label{Table_1}
	\begin{center}
		\begin{threeparttable}
			\begin{tabular}{p{2.2cm} p{1.2 cm} p{1 cm} p{1 cm} p{1cm} p{2 cm} p{2.2 cm}}
				\hline
				\toprule
				\multicolumn{1}{c} {\diagbox {\textbf{Approach}}{\textbf{Property}}} & \multicolumn{1}{c} {\textbf{Model type}} & \multicolumn{1}{c} {\textbf{Data cost}} & \multicolumn{1}{c}{\textbf{Computation complexity}} & \multicolumn{1}{c}{\textbf{Solution type}} & {\textbf{What kind of temporal correlations can be captured?}} & {\textbf{Does the temporal correlation can be clearly described?}}\\
				\hline
				\multicolumn{1}{c}{CVA} & \multicolumn{1}{c}{White-box model} & \multicolumn{1}{c}{Low} & \multicolumn{1}{c}{SVD decomposition} & \multicolumn{1}{c}{Linear} & \multicolumn{1}{c}{Short-term} & \multicolumn{1}{c}{Yes} \\

				\multicolumn{1}{c}{LSTM} & \multicolumn{1}{c}{Black-box model} & \multicolumn{1}{c}{High} & \multicolumn{1}{c}{A multi-layer RNN} & \multicolumn{1}{c}{Nonlinear} & \multicolumn{1}{c}{Long-term}  & \multicolumn{1}{c}{No} \\
				\bottomrule
			\end{tabular}
		\end{threeparttable}
	\end{center}
\end{table*}

\begin{figure*}
	\centering
	\includegraphics[scale=0.35]{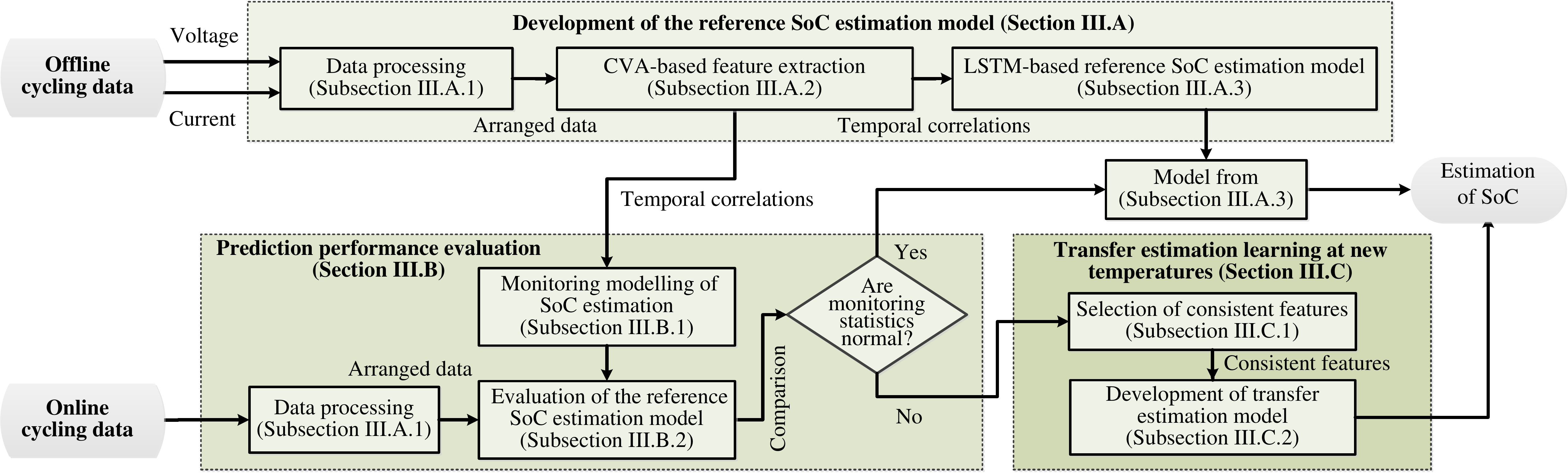}\\
	\caption{The structure and organization of the proposed method.}
	\label{MyFig2}
\end{figure*}

CVA is a dimensionality reduction method that maximizes the correlations between two datasets by finding linear combinations of high-dimensional process data \cite{Ref15}. A LSTM network is a sequence to sequence learning method, which learns how to relate an entire sequence with target state outputs in a state-space manner. To deal with the vanishing gradient problem of  traditional RNN, hidden nodes are designed as LSTM cells, where a forget gate is newly introduced and memory states are involved to remove redundant information. Due to limitation of pages, more information about these two algorithms can be found in \cite{Ref16}, \cite{Ref17}.

Although both methods are data-driven to analyze temporal dynamics, they present different working mechanisms and data requirements. Table \ref{Table_1} systematically presents the dissimilarities regarding both methods. First, CVA is a statistical ML method to find temporal correlations between variables in a linear and explainable way using latent variables. Differently, LSTM is a non-statistical ML model for long-period prediction. As a linear method, data requirements and computation complexity of CVA are much lower than that of LSTM. That is determined by the relatively simple structure of CVA. In contrast, LSTM owns a complex network, which may have multiple layers and thousands of nodes. As a result, the demands of hardware resource and amount of data to train LSTM are high.

In summary, CVA is good at revealing temporal correlations in a analytical expression with a small amount of data. And LSTM shows superiorities in finding long-period temporal prediction ability with amounts of data. To take advantages of the best of both methods for SoC estimation, an elaborate scheme is designed in this paper. The behind idea is as follows. To learn long-term temporal correlations of LiBs data, LSTM is used in its typical structure to serves as estimation model. For better estimation ability and overcoming the influence of varying temperatures, CVA is introduced as a feature extraction method and the purposes of these analytical features are threefold: 1) feeding features rather than original measurements benefits the accuracy improvement of LSTM based estimation ability; 2) extracted features allow for quantitative evaluation of the changes of estimation ability influenced by varying temperatures; 3) a prompt transfer learning works by classifying these features as consistent ones and inconsistent ones between the reference temperature and new temperatures.

\section{Methodology}
The basic structure and organization of the proposed method are shown in Fig. \ref{MyFig2} and each part is specified in the following subsections. First, a reference SoC estimation model is developed using a multi-layer LSTM network fed by latent variables extracted using CVA. Next, this reference model is evaluated by checking the variances of obtained latent variables. Finally, a TL-based rapid updating strategy is proposed if the reference model fails due to the influence of a change in temperature.

\subsection{The Reference SoC Estimation Model}
The reference SoC estimation model is developed at a reference temperature, where large amounts of data are available. LSTM is chosen for modeling by taking advantage of its ability in capturing long-period prediction. To further enhance LSTM, temporal features extracted by CVA are fed into LSTM instead of original measurements. The specific steps are:

\subsubsection{Data processing}
Considering voltage signal shows serious non-stationarity, it is improper to be directly applied in CVA and LSTM that are more suitable for stationary signals. As a multi-resolution decomposition method, wavelet analysis (WA) has been successfully applied in signal processing, which decomposes a non-stationary signal into a series of stationary components. The core idea behind WA is that components of signal at corresponding frequencies will be amplified and easy to recognize through a series of wavelet functions that are finite energy and smooth. Morlet et al. \cite{Ref18} proposed the discrete wavelet analysis (DWA) method, in which a signal was decomposed into an average value and a series details. Decomposition of current signal $\mathbf c_n$ and voltage signal $\mathbf v_n$ in discharge cycle $n$ using DWA is given below,
\begin{equation}
\begin{array}{l}
{\mathbf{c}_{n}=\mathbf{c}_{n,a}+\sum_{j=1}^{J_{c}} \mathbf{c}_{n,j}} \\ [1mm]
{\mathbf{v}_{n}=\mathbf{v}_{n,a}+\sum_{j=1}^{J_{v}} \mathbf{v}_{n,j}}
\end{array}
\end{equation}
where $\mathbf c_{n,a}$ and $\mathbf v_{n,a}$ are the average part reflecting the low-frequency information of current and voltage in the $n^{th}$ cycle, respectively; $\mathbf c_{n,j}$ and $\mathbf v_{n,j}$ are the $j^{th}$ detail reflecting high-frequency information of the $n^{th}$ current and voltage, respectively; $J_c$ and $J_v$ are the number of details of current and voltage, respectively.

The original two-dimensional data matrix $\mathbf X_n$ is extended into a new matrix with dimensions $K_n \times J_x$, in which $J_x=J_c+J_v+2$ and $K_n$ is the number of samplings of the $n^{th}$ cycle. For brevity, the new matrix is still denoted as $\mathbf X_n$.

\begin{figure}[tb]
\centering
\includegraphics[scale=0.55]{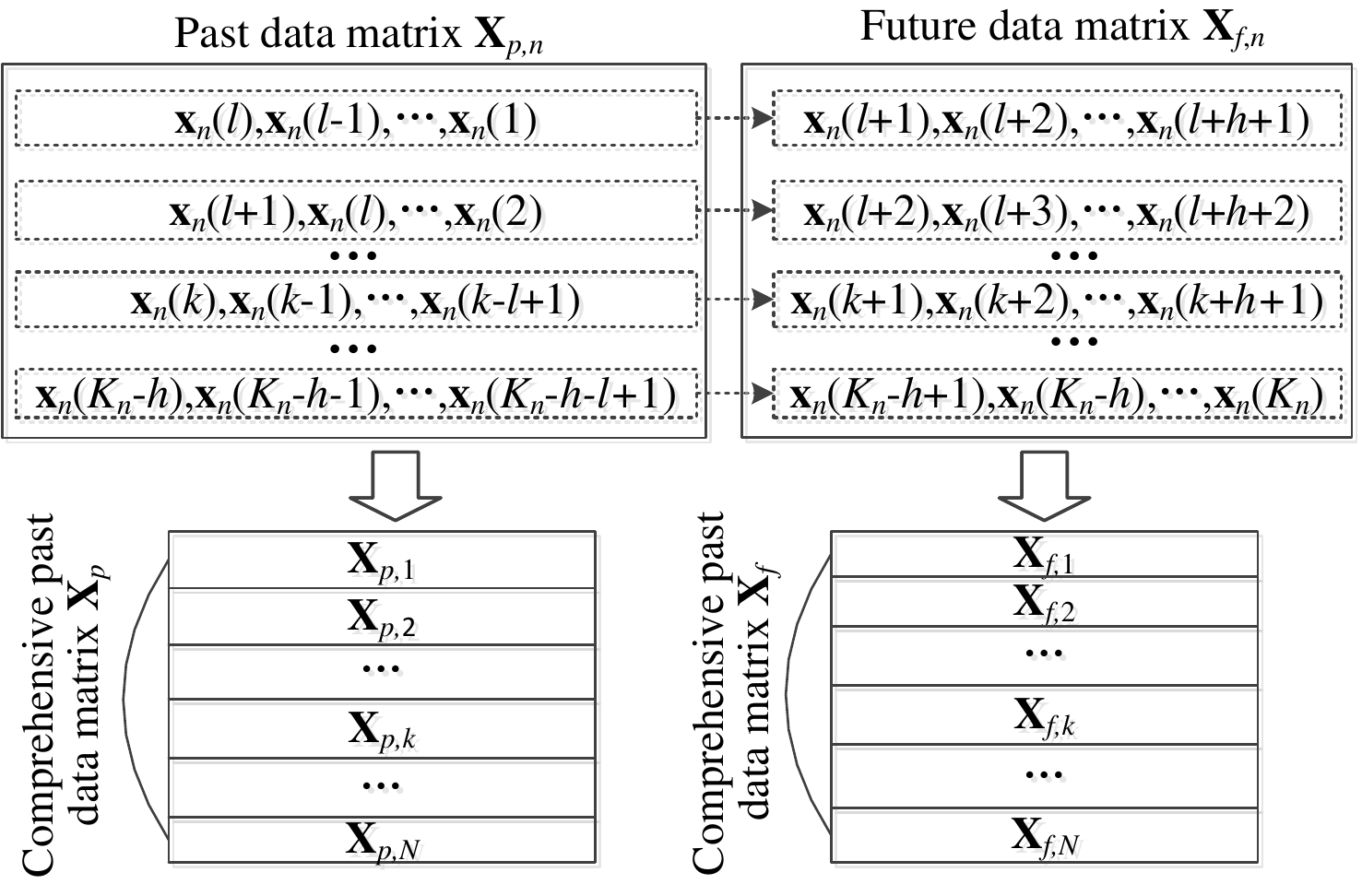}\\
\caption{Schematic strategy of the past data and the future data matrixes.}
\label{MyFig3}
\end{figure}

\subsubsection{CVA-based feature engineering} Although advanced deep learning networks are capable of developing estimation models with raw data,  feature engineering plays a vital role in further boosting the performance of data-driven models [19]-[21]. For instance, voltage falloff, a health indicator to average the discharging voltage with an equivalent interval of the whole discharging cycle, has successfully reduced the battery RUL estimation error [19], [20]. In this article, CVA-based feature engineering is preferred due to its ability to capture temporal correlation from time-series.

To relate the temporal property, past data vector $\mathbf x_{p,n}(k)=[\mathbf x_n(k-1), \mathbf x_n(k-2),..., \mathbf x_n(k-l)]$ and future data vector $\mathbf x_{f,n}(k)=[\mathbf x_n(k), \mathbf x_n (k+1),..., \mathbf x_n (k+h-1)]$ are formed at each sampling time, where $\mathbf x_n (k)$ is the $k^{th}$ row of $\mathbf X_n$; $p$ and $f$ indicate past information and future information, respectively. The parameters $l$ and $h$ relate to how many past data and future data will be used for modeling, which meets well with the purpose of the autocorrelation function. Thus, the specific values of these two parameters are calculated using the root summed squares of all variables from the normal data against certain confidence bound ($\pm 5\%$ here). Usually, these two parameters are given with the same value.

Then past data matrix $\mathbf X_{p,n}$ and future data matrix $\mathbf X_{f,n}$ are gained as ${\mathbf{X}_{p,n}=[\mathbf{x}_{p,n}(l), \mathbf{x}_{p,n}(l+1), ..., \mathbf{x}_{p,n}(K_n-h)]}$ and ${\mathbf{X}_{f,n}=[\mathbf{x}_{f,n}(l+1), \mathbf{x}_{f,n}(l+2), ..., \mathbf{x}_{f,n}(K_n-h+1)]}$. Fig. 3 illustrates the data arrangement procedure. Further, comprehensive past and future matrixes $\mathbf X_p$ and $\mathbf X_f$ are formed by putting $\mathbf X_{p,n}$ and $\mathbf X_{f,n}$ in each cycle together along variable-wise direction, respectively. With $\mathbf X_p$ and $\mathbf X_f$, an eigenvalue decomposition is performed below,
\begin{equation}
\mathbf \Sigma_{p p}^{-1/2} \mathbf \Sigma_{p f} \mathbf \Sigma_{f f}^{-1/2}=\mathbf{U}_{x} \boldsymbol{\Sigma}_{x} \mathbf{V}_{x}
\end{equation}
where $\mathbf \Sigma_{p p}$ and $\mathbf \Sigma_{f f}$ are the covariance matrixes of $\mathbf X_p$ and $\mathbf X_f$, respectively; $\mathbf \Sigma_{p f}$ is the cross-covariance matrix between $\mathbf X_p$ and $\mathbf X_f$; $\mathbf \Sigma_{x}$ is a diagonal matrix of nonnegative singular values; $\mathbf U_x$ and $\mathbf V_x$ are right-singular matrix and left-singular matrix, respectively.

In this way, CVs $\mathbf Z_x$ in $\mathbf X_p$ are calculated as linear combinations of $\mathbf X_p$ by retaining all features as follows,
\begin{equation}
\mathbf{Z}_{x}=\mathbf{J}_{x} \mathbf{X}_{p}=\mathbf{V}_{x}^{\mathrm{T}} \boldsymbol{\Sigma}_{p p}^{-1/2} \mathbf{X}_{p}
\end{equation}

Similarly, CVs $\mathbf Z_y$ in $\mathbf X_f$ are calculated as $\mathbf{Z}_{y}=\mathbf{J}_{y} \mathbf{X}_{f}=\mathbf{U}_{x}^{\mathrm{T}} \boldsymbol{\Sigma}_{f f}^{-1/2} \mathbf{X}_{f}$. It should be noted that both $\mathbf X_p$ and $\mathbf X_f$ should be normalized to zero mean and unit variance before being applied in Eq. (2).

\subsubsection{LSTM-based reference SoC estimation model}
A long-period prediction ability is achieved through a multi-layer LSTM network, which consists of one input layer, several LSTM hidden layers, and an output layer. Taking $\mathbf Z_x$ as inputs and the real values of SoC $\mathbf Y_x$ as ground truth, the reference SoC estimation model $\Gamma (\cdot)$ achieves sequential to sequential learning through minimizing the loss function $L$ between real values and estimations below,
\begin{equation}
\min \limits_{\Gamma(\cdot)} L=\sum_{k=1}^{K_{z}}\left(y(k)-\Gamma\left(\mathbf{z}_{x}(k)\right)\right)^{2}
\end{equation}
where $\mathbf z_x(k)$ is the $k^{th}$ row of $\mathbf Z_x$, $y(k)$ is the real values in $\mathbf Y_x$, and $K_z$ is the total number of samples in $\mathbf Z_x$.

In summary, the inputs of LSTM in reference SoC model are temporal feature $\mathbf Z_x$ and corresponding real SoC values as ground truth, and the well-trained model outputs predictions of SoC for new cycling data. The benefits of developing the hybrid architecture of CVA-based feature engineering and LSTM are twofold. On the one hand, the temporal correlations that contribute to estimation ability are specified. On the other hand, the advantages of the long-term learning ability of LSTM can be retained. It is worth noticing that interpretable insights on the correlation of individual features into the target output could be achieved by adopting interpretable LSTM \cite{Ref32}. During training, Adam optimization \cite{Ref22} is employed to update the weights and biases.

\subsection{Prediction Performance Evaluation}
Having a close relationship with estimation ability, temporal correlation is used as an indicator to evaluate SoC estimation performance by checking its variability. Therefore, $\mathbf Z_x$ is used to develop monitoring model based on multivariate statistical process analysis.
\subsubsection{Monitoring modelling of SoC estimation}
Diagonal elements of $\mathbf \Sigma_{x} = \operatorname{diag}(\alpha_1,\alpha_2,...,\alpha_{J_x l})$ in Eq. (2) reflect the correlation of pair-wise CVs in $\mathbf Z_x$ and $\mathbf Z_y$. For instance, $\alpha_1=\mathbf Z_{x,1}^T \mathbf Z_{y,1}$, where $\mathbf Z_{x,1}$ and $\mathbf Z_{y,1}$ are the first column of $\mathbf Z_x$ and $\mathbf Z_y$, respectively. According to the amplitude of correlation, $\mathbf Z_x$ are classified into two categories: system canonical variates (SCVs) and the residual variates (RVs). SCVs reflect main CVs with large correlations. And CVs with small correlations are grouped into RVs. Given there are $R$ SCVs, $\textbf Z_x$ can be decomposed as follows,
\begin{equation}
\begin{array}{l}
{\mathbf{Z}_{s}=\mathbf{J}_{s} \mathbf{X}_{p}=\mathbf{V}_{s}^{\mathrm{T}} \mathbf{\Sigma}_{p p}^{-1/2} \mathbf{X}_{p}} \\ [1mm]
{\mathbf{Z}_{r}=\mathbf{J}_{r} \mathbf{X}_{p}=(\mathbf{I}-\mathbf{V}_{s} \mathbf{V}_{s}^{\mathrm{T}}) \mathbf \Sigma_{p p}^{-1/2} \mathbf{X}_{p}}
\end{array}
\end{equation}
where $\mathbf Z_s$ contains SCVs, $\mathbf Z_r$ consists of RVs, $\mathbf V_s$ is the first $R$ columns of $\mathbf V_x$ in Eq. (2), and $\mathbf I$ is an identity matrix with dimension $J_xl$. Besides, $s$ and $r$ refer to the system subspace spanned by SCVs and the residual subspace spanned by RVs, respectively. The parameter $R$ stands for the number of dominant singular values, which is determined by finding a ``knee" point \cite{Ref23} from the singular matrix $\mathbf \Sigma_{x}$ calculated from Eq. (2).

To quantitatively measure the variance of SCVs and RVs, two monitoring statistics, i.e., indicators, are constructed using Hotelling's $T^2$ and squared prediction error ($SPE$) \cite{Ref24}. At sampling time $k$, they are calculated as follows,
\begin{equation}
\begin{array}{l}
{T^{2}(k)=\sum_{i=1}^{R} z_{s,i}(k)^{\mathrm{T}} z_{s,i}(k)} \\ [1mm]
{SPE(k)=\sum_{i=1}^{J_{x} l} z_{r,i}(k)^{\mathrm{T}} z_{r,i}(k)}
\end{array}
\end{equation}
where $z_{s,i}(k)$ is the $k^{th}$ row and the $i^{th}$ column in $\mathbf Z_s$, and $z_{r,i}(k)$ is the $k^{th}$ row and the $i^{th}$ column in $\mathbf Z_r$.

After checking the normality of $\mathbf Z_s$ and $\mathbf Z_r$ \cite{Ref25}, corresponding statistical control limits can be derived. If $\mathbf Z_s$ and $\mathbf Z_r$ follow Gaussian distribution, $\chi^2$ and $F$ distribution can be used to derive the statistical control limits $CL_s$ and $CL_r$ \cite{Ref26} with a given significance level, which is specified as 0.95 here. That is, the control limits will cover 95$\%$ normal samples. Otherwise, kernel density estimation \cite{Ref27} is suggested to estimate the probability distribution of $T^2$ and $SPE$ when $\mathbf Z_s$ and $\mathbf Z_r$ are non-Gaussian distribution, and then corresponding control limits can be calculated.

\subsubsection{Evaluation of the reference SoC estimation model}
For online monitoring, a data vector $\mathbf x_{new}(J_xl \times 1)$ is collected, which meets with the required window length of DWA. First, perform DWA on $\mathbf x_{new}$ according to Eq. (1) to obtain multi-solution information and then obtain the past data vector. Next, the normalized past data vector, which is still denoted as $\mathbf x_{new}$, is projected on the system subspace $\mathbf J_s$ and the residual subspace ${\mathbf J}_{r}$ given in Eq. (5) to get SCVs $\mathbf z_{new,s}=\mathbf J_s \mathbf x_{new,s}$ and the residuals $\mathbf z_{new,r}=\mathbf J_r \mathbf x_{new,r}$, respectively, in which $s$ and $r$ have the same meaning as that in Eq. (5). After that, the online monitoring statistics of the system subspace and the residual subspace are computed below,
\begin{equation}
\begin{aligned}
t^{2} &=\sum_{i=1}^{R} z_{new,s,i}^{\mathrm{T}} z_{new,s,i} \\
spe &=\sum_{i=1}^{J_xl} z_{new,r,i}^{\mathrm{T}} z_{new,r,i}
\end{aligned}
\end{equation}
where $z_{new,s,i}$ and $z_{new,r,i}$ are the $i^{th}$ element of $\mathbf z_{new,s}$ and $\mathbf z_{new,r}$, respectively.

\begin{figure}[htb]
	\centering
	\includegraphics[scale=0.75]{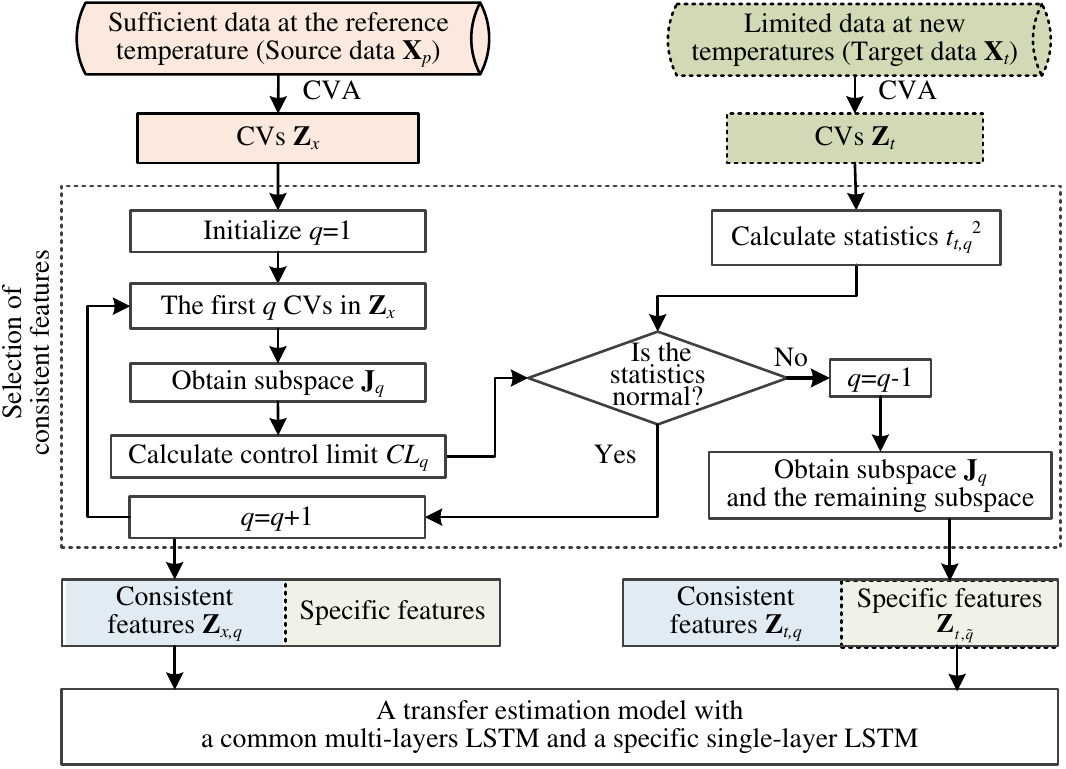}\\
	\caption{Flow chart of the transfer estimation learning at new temperatures.}
	\label{MyFig3}
\end{figure}

By sliding a moving window with fixed window length of DWA, $\mathbf x_{new}$ is updated with time evolution and monitoring of reference SoC model will proceed. By continuous comparing the online monitoring statistics with their control limits, performance of the reference SoC estimation model is judged according to the following rules:

Case I: If both online monitoring statistics are below their control limits, it indicates that SCVs and RVs both stay normal and estimation ability is not influenced by changes in temperature. Thus, the reference SoC estimation model trained in Eq. (4) can be directly adopted for estimation,
\begin{equation}
\hat{y}_{new}=\Gamma (\mathbf{z}_{new})
\end{equation}
where $\mathbf{z}_{new}$ is the collection of $\mathbf{z}_{new,s}$ and $\mathbf{z}_{new,r}$.

Case II: If at least one of online monitoring statistics is above its control limit in continuous three samplings, it means that partial CVs have been changed due to temperature variations. In this case, the current estimation relationship is improper and the reference SoC estimation model needs to be updated through TL, which will be shown in Subsection III.C.

\subsection{Transfer Estimation Learning at New Temperatures}
Once monitoring statistics are abnormal, it indicates that the reference SoC estimation model is no longer feasible to be used. Indeed, though temporal correlations may be dissimilar when temperature varies, consistent temporal correlations, which keep the same at different temperatures, may exist due to the fundamental nature of the electrochemical process in LiBs. Inspired by this, the basic idea of the proposed transfer estimation model is to inherit the common prediction ability from the source task meanwhile exploiting the ability of specific prediction in the target task.

The modeling procedures at the reference and the new temperatures are regarded as the source and the target task. Naturally, this raises two key issues when transfer process is triggered: 1) what to transfer, and 2) how to transfer. Fig. \ref{MyFig3} sketches the essence of the proposed transfer strategy, which includes two corresponding parts. The selection of consistent features from the reference SoC estimation model solves the first problem. The proposed transfer estimation model addresses the second one by combining consistent features from the reference temperature and the specific features at new temperatures.

\begin{figure*}[!ht]
\centering
\includegraphics[scale=0.4]{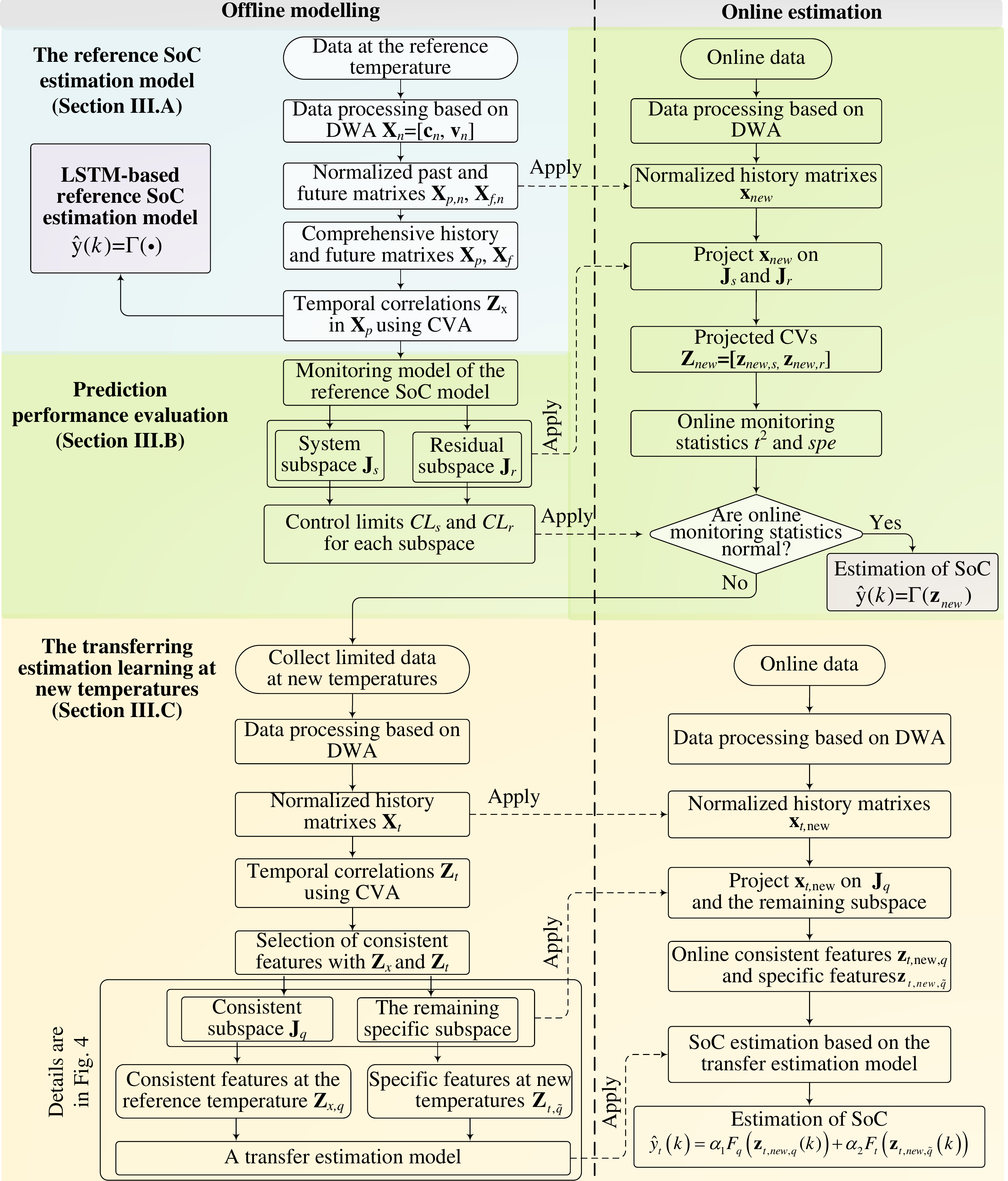}\\
\begin{center}
\caption{Detailed illustration of the proposed method.}
\end{center}
\label{MyFig5}
\vspace{-0.5cm}
\end{figure*}

Let $\mathbf X_t$ a complete discharge cycle with current and voltage collected at new temperatures. $\mathbf X_t$ is decomposed using DWA and the past data matrix is arranged. The past data matrix is normalized with its own information and still denoted as $\mathbf X_t$. Correspondingly, the real values of SoC at new temperatures are denoted as $\mathbf Y_t$. To investigate the temporal dynamics of $\mathbf X_t$, CVA is performed on $\mathbf X_t$ in an analogous way to Eq. (2). CVs $\mathbf Z_t$ in target temperature are calculated by retaining all features below,
\begin{equation}
\mathbf{Z}_{t}=\mathbf{J}_{t} \mathbf{X}_{t}=\mathbf{V}_{t}^{\mathrm{T}} \mathbf {\Sigma}_{t t}^{-1/2} \mathbf{X}_{t}
\end{equation}
where $\mathbf J_t$ is projection matrix, $\mathbf V_t$ is left-singular matrix, and $\mathbf {\Sigma}_{t t}$ is covariance matrix of $\mathbf X_t$.

\begin{algorithm}[!ht]
	\scriptsize
	\caption{Selection of consistent features}
	\label{alg1}
	\begin{algorithmic}[1]
		\STATE Input: $\textbf Z_x$ and $\textbf Z_t$
		\STATE Initialize $q$ = 1
		\FOR{$q = 1, 2, ..., J_xl$}
		\STATE Retain the first $q$ CVs in $\textbf Z_x$ as $\textbf Z_{x,q}$
		\STATE Calculate $\textbf J_q$ and control limit $CL_q$ according to Eqs. (5) and (6) from $\textbf Z_{x,q}$
		\STATE Denote the first $q$ features in $\textbf Z_t$ as $\textbf Z_{t,q}$
		\STATE Calculate statistics $t_{t,q}^2$ from $\textbf Z_{t,q}$ at each sampling time according to Eq. (7)
		\IF    {$t_{t,q}^2 \leqslant CL_q$ holds for all sampling times}
		\STATE $q=q+1$
		\ELSE  [$t_{t,q}^2> CL_q$ holds at continuous three sampling times]
		\STATE $q=q-1$
		\ENDIF
		\ENDFOR
	\end{algorithmic}
\end{algorithm}

\subsubsection{Selection of consistent features}
Assuming the first $q$ features in $\mathbf Z_t$ are consistent with that of $\mathbf Z_x$, the monitoring statistics will be under the corresponding control limits calculated from the reference temperature. As a result, the proper value of $q$ can be determined by checking the variations of consistent features in $\mathbf Z_t$ with respect to the control limits derived from $\mathbf Z_x$. To select consistent features, a sequential selection strategy is given in {\textbf {Algorithm 1}}.

With the above procedures, the number of consistent features $q$ between the reference temperatures and target temperatures is determined. After that, consistent features $\mathbf Z_{x,q}$ in reference temperature are derived by separating the first $q$ rows in $\mathbf Z_x$. The consistent features in new temperatures is obtained as $\mathbf Z_{t,q}$, which is the collection of the first $q$ feature in $\textbf Z_t$. Correspondingly, the specific features $\mathbf Z_{t, \widetilde q}$ in new temperatures is computed by removing $\mathbf Z_{t,q}$ from $\textbf Z_t$.

\subsubsection{Development of transfer estimation model}
The proposed transfer estimation model consists of two independent LSTM networks. The first one is a multi-layer LSTM network that aims to learn shared prediction ability by training data pair $\{\mathbf Z_{x,q}, \mathbf Y_{sour}\}$, in which $\mathbf Y_{sour}$ is the real values of SoC in source task, i.e., the reference temperature. The other one is a single-layer LSTM network, which will be trained using a small amount of data pair $\{\mathbf Z_{t,\widetilde q}, \mathbf Y_t \}$, in which $\mathbf Y_t$ is the real values of SoC in target task (the new temperature). These two networks are independently designed for rapid TL purpose. The shared prediction ability $F_q(\cdot)$ is derived by minimizing estimation errors below,
\begin{equation}
\begin{array}{l}
{\min \limits_{F_{q}} \sum {({\hat{\mathbf Y}_{sour}}(k)-\mathbf{Y}_{sour}(k))^{2}}} \\
{{s.t.} \hat{\mathbf{Y}}_{sour}(k)=F_{q}(\mathbf{Z}_{x,q}(k))}
\end{array}
\end{equation}
where $\hat {\mathbf Y}_{sour}(k)$ is the estimation of $\mathbf Y_{sour}(k)$.

With $F_q(\cdot)$, the target task is represented as follows,
\begin{equation}
\begin{array}{l}
{\min \limits_{F_{t}} \sum (\hat{\mathbf{Y}}_{t}(k)-\mathbf{Y}_{t}(k))^{2}} \\
{{s.t.} \hat{\mathbf{Y}}_{t}(k)=\alpha_{1,k} F_{q}(\mathbf{Z}_{x,q}(k))+\alpha_{2,k} F_{t}(\mathbf{Z}_{t, \widetilde q}(k))}
\end{array}
\end{equation}
where $\hat {\mathbf Y}_t(k)$ is the estimation of $\mathbf Y_t(k)$; $\mathbf Z_{x,q}(k)$ and $\textbf Z_{t,\widetilde q}(k)$ are the $k^{th}$ row of $\mathbf Z_{x,q}$ and $\mathbf Z_{t,\widetilde q}$, respectively; $F_t(\cdot)$ is the specific prediction ability with respect to target temperature; $\alpha_{1,k}$ and $\alpha_{2,k}$ are real-time adjusting factors that both are initialized to be 0.5.

To balance the shared prediction ability and the specific prediction ability in Eq. (11), a strategy proposed in \cite{Ref28} is employed to update adjusting factors below,
\begin{equation}
\begin{array}{l}
{\alpha_{1,k}=\frac{{\alpha_{1,k} \Psi_{k}}(F_{q}(\mathbf{Z}_{x,q}(k)))}{\alpha_{1,k} \Psi_{k}(F_{q}(\mathbf{Z}_{x,q}(k)))+\alpha_{2, k} \Psi_{k}(F_{t}(\mathbf{Z}_{t,\widetilde q}(k)))}} \\ [2mm]
{\alpha_{2,k}=\frac{\alpha_{2,k} \Psi_{k}(F_{t}(\mathbf{Z}_{t,\widetilde q}(k)))}{\alpha_{1,k} \Psi_{k}(F_{q}(\mathbf{Z}_{x,q}(k)))+\alpha_{2,k} \Psi_{k}(F_{t}(\mathbf{Z}_{t,\widetilde q}(k)))}}
\end{array}
\end{equation}
where $\Psi_{k}(f(\cdot))=\exp (-\eta(f(\cdot)-y(k))^{2})$, $y(k)$ is the corresponding real value, and $\eta \in [0,1]$ is a coefficient.

From Eqs. (11) and (12), input data of the transfer estimation model include two parts. One is the consistent features $\mathbf Z_{x,q}$ with the corresponding real values of SoC $\mathbf Y_{sour}$ in the reference temperature as the label, and the other one is the specific features $\mathbf Z_{t,\widetilde q}$ with the corresponding real values $\mathbf Y_t$ in the new temperature as labels. Once the SoC transfer estimation model is well-trained, online estimations can be achieved for SoC values at the target temperature.
Here, the new sample at the $k^{th}$ sampling time is denoted as $\mathbf x_{t,new}(k)$, and then past data vector is arranged and normalized. After that, $\mathbf x_{t,new}(k)$ is projected on the subspace $\mathbf J_q$ expanded by consistent features and the remaining subspace. The consistent features and the specific features are obtained as $\mathbf z_{t,new,q}(k)=\mathbf J_{q} \mathbf x_{t,new}(k)$ and $\mathbf z_{t,new,\widetilde q}(k)=(\mathbf I - \mathbf J_{q}) \mathbf x_{t,new}(k)$. With $F_q(\cdot)$ and $F_t(\cdot)$, online estimation $\hat{y}_{t}(k)$ in new temperature is achieved below,
\begin{equation}
\hat{y}_{t}(k)=\alpha_{1}F_{q}(\mathbf z_{t,new,q}(k))+\alpha_{2}F_{t}(\mathbf z_{t,new,\widetilde q}(k))
\end{equation}
where $\alpha_1$ and $\alpha_2$ are the final values of adjusting factors.

For better understanding, Fig. 5 summarizes the main steps of the proposed method from off-line modelling and online estimation.

\begin{figure}[!ht]
\centering
\subfigure[]{
\begin{minipage}[t]{0.5\linewidth}
\centering
\includegraphics[width=1.8in]{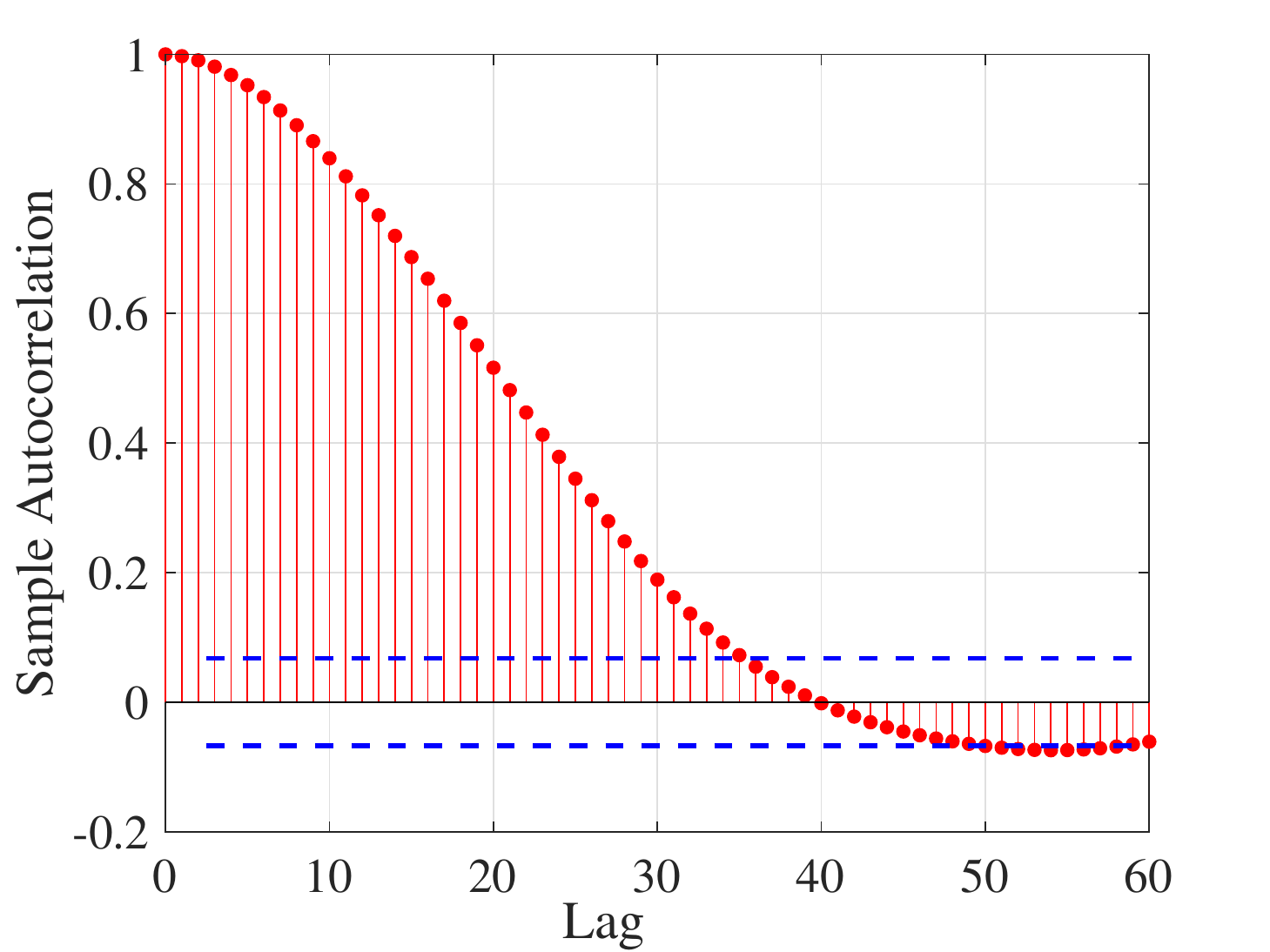}
\end{minipage}%
}%
\subfigure[]{
\begin{minipage}[t]{0.5\linewidth}
\centering
\includegraphics[width=1.8in]{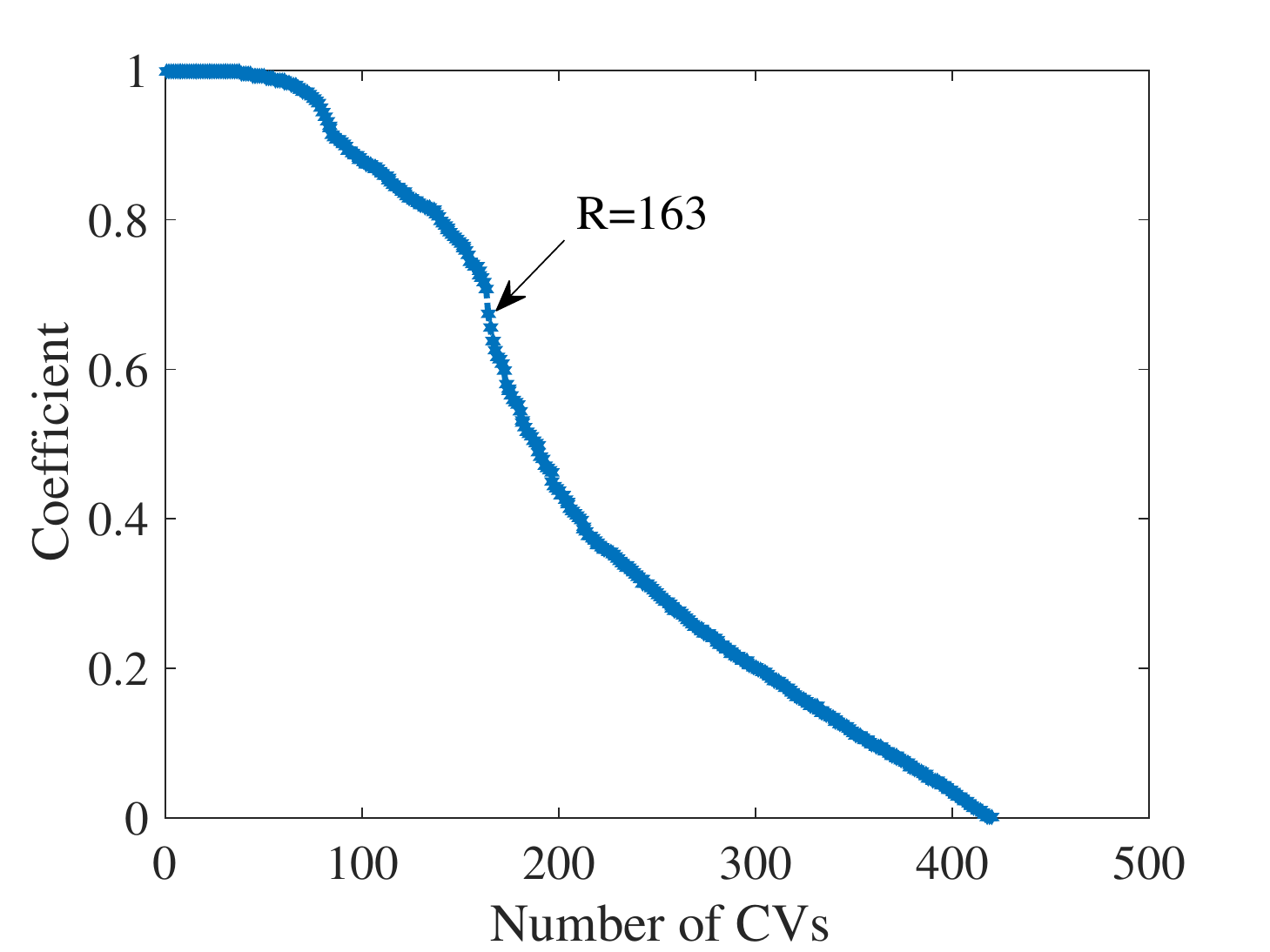}
\end{minipage}
}
\centering
\caption{Selection of propoer values for tuanble parameters at the reference temperature 10$^\circ$C with respect to (a) time lags $l$ and $h$ for reference estimation model and (b) the number of retained CVs $R$ for prediction performance evaluation model.}
\vspace{-0.5cm}
\end{figure}

\section{Results and discussions}
In this section, the proposed method is verified through a widely used LiB benchmark, which is provided by Dr. Kollmeyer \cite{Ref14}. It provides a series of drive cycles simulating an electric Ford F150 truck powered by a 35-kWh stack of 2.9 Ah Panasonic 18650PF cells (containing NCA oxide cathodes) at five discrete ambient temperatures (25$^\circ$C, 10$^\circ$C, 0$^\circ$C, -10$^\circ$C, and -20$^\circ$C). Ten drive cycles are available at each temperature, and each cycling data includes four variables, including current, voltage, battery surface temperature, and amp-hours. The first two variables are employed in the proposed method as input variables. The last variable amp-hours is measured to infer the real value of SoC. And the third variable is not used in our model to leave temperature as an external condition. All measurements are recorded at a frequency of 1Hz.

For fair comparisons, all simulations and data analysis are conducted on a desktop equipped with 16 processors of Intel Xeon E5-2620 v4 (20 MB cache, up to 2.10 GHz) and a multi-graphics processor unit of NVIDIA GeForce GTX 1080Ti at 11 GB. Programming software is Matlab with the version of 2019a. To evaluate the prediction accuracy, indices root mean square error ($RMSE$) and mean absolute error ($MAE$) are comprehensively used, which are given as follows,
\begin{equation}
\begin{array}{l}
{RMSE=\sqrt{\frac{1}{K} \sum_{k=1}^{K}(y_{t}(k)-\hat{y}_{t}(k))^{2}}} \\ [1mm]
{MAE=\frac{1}{K} \sum_{k=1}^{K}|y_{t}(k)-\hat{y}_{t}(k)|}
\end{array}
\end{equation}
where $\hat y_t(k)$ and $y_t(k)$ have the same meaning as that in Eq. (13), and $K$ is the number of sampling times in a testing cycle.

For all metrics, a smaller value indicates a better prediction accuracy [30], [31]. It should be noted that these indices are calculated with a complete cycling data.

\subsection{SoC Estimation Comparisons at Fixed Temperatures}
This part focuses on illustrating estimation performance of the proposed model at fixed temperatures. Five reference SoC models are developed corresponding to each temperature, where ten discharging cycles are available. Here, the first nine drive cycles are used for training, including one cycling data for validation. And the last cycling data is employed for testing. Considering the experiments are mainly conducted on 10$^\circ$C in \cite{Ref8}, comparisons are first completed at this temperature. Performing DWA on the current signal and the voltage signal of training data, each signal is decomposed into six components through retaining five details. Further, by conducting CVA on the decomposed data, a series of CVs are obtained according to Eqs. (2) and (3). In CVA, two crucial parameters $l$ and $h$ need to be determined. According to rules given in Section III.A, values of these two parameters are calculated using the root summed squares of all variables from the normal data against a certain confidence bound ($\pm 5 \%$ here). Therefore, both $l$ and $h$ are set to be 36 as shown in Fig. 6(a), where it is observed that the value steps into the defined confidence region \cite{Ref31}.

\begin{table*}[ht]
	\tiny
	\renewcommand{\arraystretch}{1.2}
	\caption{Comparisons between the proposed method and LSTM at five discrete temperatures concerning accuracy and network configuration.}
	\label{Table_2}
	\begin{center}
		\begin{threeparttable}
			\begin{tabular}{c c c c| c c c| c}
				\toprule
				\multicolumn{1}{c}{ \multirow{2}*{\textbf{Temp. ($^\circ$C)}}} & \multicolumn{1}{c}{\multirow{2}*{\textbf{Method}}} & \multicolumn{2}{c|}{\textbf{Indices}} & \multicolumn{3}{c|}{\textbf{Network configurations}} & \multicolumn{1}{c}{ \multirow{2}*{\textbf{Values of tunable parameters}}}\\
				\cline{3-7}
				\multicolumn{1}{c}{} & \multicolumn{1}{c}{} & $RMSE$ & $MAE$ & Iteration epochs & Training time & Network\\
				\hline
				\multirow{3}*{25} & The proposed method & 2.74 & 2.27 & 2000 & 105min34sec & L(50, 100)N(100) & $l=34,h=34,R=160$ \\
				& LSTM-based method & 5.46 & 4.29 &	15000 & 127min38sec & L(400)N(100) & - \\
				& Accuracy improvement & 49.82$\%$ & 47.09$\%$ & - & - & - & - \\
				\cline{1-8}
				\multirow{3}*{10} & The proposed method & 0.58& 0.46 & 2000 & 55min30sec &  L(50, 100)N(100) & $l=36,h=36,R=163$ \\
				& LSTM-based method & 1.41 & 1.12 & 15000 & 155min10sec & L(500)N(100) & -  \\
				& Accuracy improvement & 58.87$\%$ & 58.93$\%$ & - & - & - & - \\
				\cline{1-8}
				\multirow{3}*{0} & The proposed method & 1.91 &	1.70 & 2000 & 63min0sec &  L(50, 100)N(100)   & $l=40,h=40,R=165$  \\
				& LSTM-based method & 2.15 & 1.84 & 15000 & 135min3sec & L(500)N(100) & - \\
				& Accuracy improvement & 11.16$\%$ & 7.61$\%$ & - &	- & - & - \\
				\cline{1-8}
				\multirow{3}*{-10} & The proposed method &	2.89 & 2.17 & 3500 & 88min15sec &  L(150, 250 )N(100) & $l=35,h=35,R=155$ \\
				& LSTM-based method & 4.00 & 3.15 & 15000 & 116min45sec & L(400)N(100) & -   \\
				& Accuracy improvement & 27.75$\%$ & 31.11$\%$ & - & - & - & - \\
				\cline{1-8}
				\multirow{3}*{-20} & The proposed method & 4.94 & 3.95 & 2000 & 35min9sec &  L(50, 100)N(100) & $l=36,h=36,R=150$ \\
				& LSTM-based method & 6.53 & 5.14 & 15000	& 116min20sec & L(400)N(100) & - \\
				& Accuracy improvement & 24.35$\%$ & 23.15$\%$ & - & - & - & - \\
				\bottomrule
			\end{tabular}
			\begin{tablenotes}
				\footnotesize
				\item[-] denotes null. 
			\end{tablenotes}
		\end{threeparttable}
	\end{center}
	\vspace{-0.8cm}
\end{table*}

\begin{figure}
	\centering
	\includegraphics[scale=0.55]{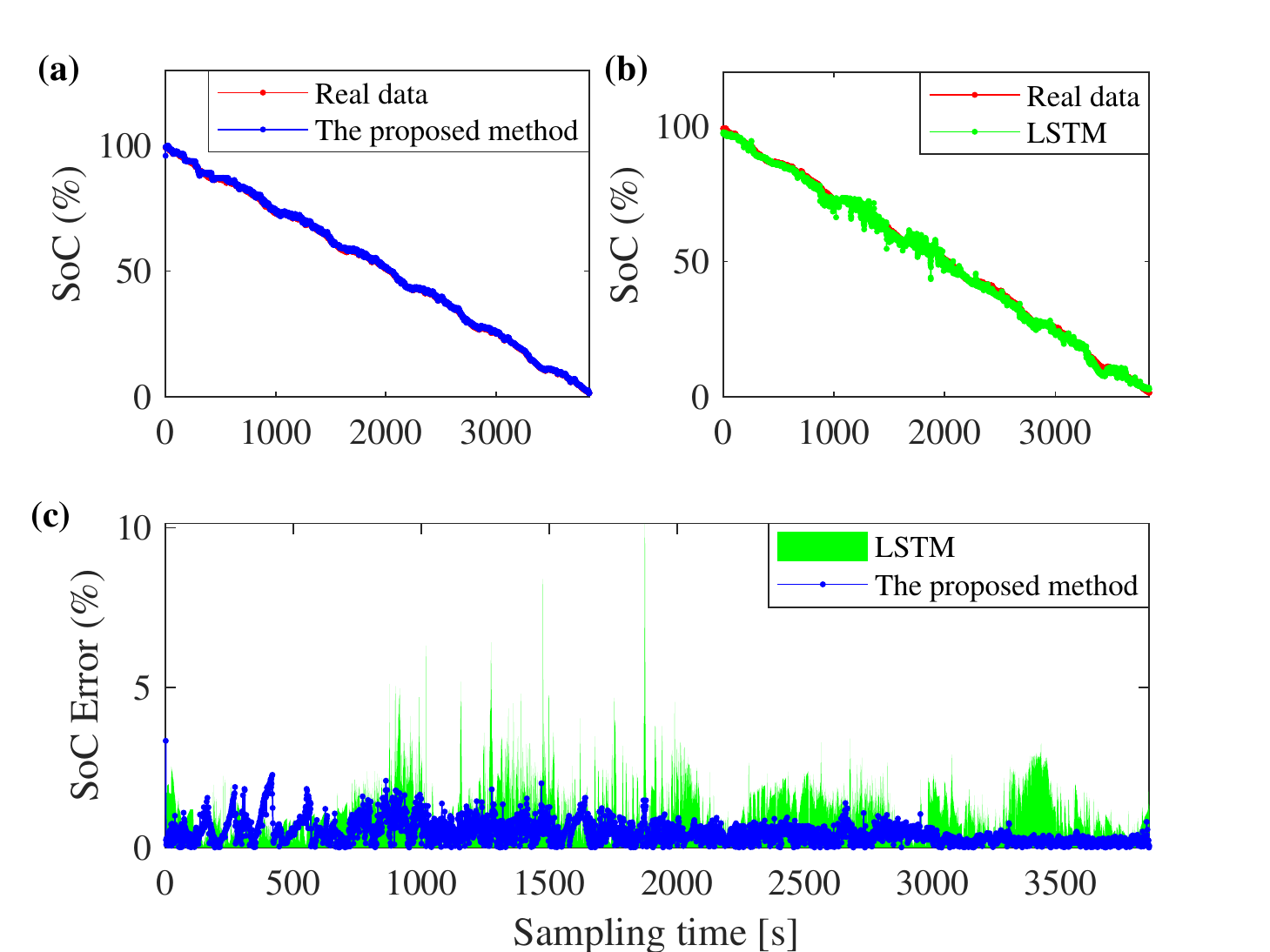}
	\caption{Prediction result comparisons between the proposed method and the LSTM-based method at 10$^\circ$C using the testing data.}
	\label{MyFig7}
	\vspace{-0.7cm}
\end{figure}

Inputting all CVs into the LSTM network with configurations given in Table II, Figs. 7(a) and 7(b) plot the predictions of SoC at 10$^\circ$C based on the proposed method and traditional LSTM-based method using testing discharge cycle, respectively. Fig. 7(c) further compares the absolute value of estimation error for both methods to give a better demonstration. Moreover, Table II summarizes the results at the other four temperatures with defined indices and specific values of tunable parameters. Values of each index at all temperatures have been greatly reduced using the proposed method. For 10$^\circ$C, index $RMSE$ is reduced by 58.87$\%$ from 1.41 (LSTM-based method) to 0.58 (the proposed method). The value is calculated through the ratio between reduced error and the original error, i.e., (1.41-0.58)/1.41. Accuracy improvements with respect to $RMSE$ are gained as 49.82$\%$, 11.16$\%$, 27.75$\%$, and 24.35$\%$ at 25$^\circ$C, 0$^\circ$C, -10$^\circ$C, and -20$^\circ$C, respectively. Therefore, prediction accuracy has been greatly improved and it proves that utilization of temporal dynamics enhances the precision of the results. Similar conclusions can be drawn with respect to index $MAE$.

\begin{table*}[ht]
	\tiny
	\renewcommand{\arraystretch}{1.2}
    	\centering
	\caption{Comparison of estimation errors using the proposed method with and without transfer learning.}
	\label{Table_3}
	\begin{center}
		\begin{threeparttable}
			\begin{tabular}{c c c c c| c c c| c}
				\toprule
				\multicolumn{1}{c}{\multirow{2}*{\textbf{Transferring case}}} & \multicolumn{1}{c}{\multirow{2}*{\textbf{Method}}} & \multicolumn{3}{c|}{\textbf{Indices}} & \multicolumn{3}{c|}{\multirow{1}*{\textbf{Transfer network configuration}}} & \multicolumn{1}{c}{\multirow{2}*{\textbf{Values for parameters}}} \\
				\cline{3-8}
				{} & {} & {$RMSE$} & {$MAE$} & {$H(\%)$} & {Iteration epochs} & {Training time} &{Network}\\
				\hline
				\multirow{3}*{From 10$^\circ$C to 25$^\circ$C} & {Model with TL\tnote{1}} & {6.93} & {5.73} & {86.44} & {100} & {1min45sec} & L(50)N(100) & {$q=200, \alpha_1=0.54, \alpha_2=0.46$}\\
				& {Model without TL\tnote{2}} & {33.77} & {29.37} & {-}  & {-} & {-}  & {-}  & {-} \\
				& {Accuracy improvement} & {79.48$\%$} & {80.49$\%$} & {-} & {-} & {-} & {-}  & {-} \\
				\cline{1-9}
				\multirow{3}*{From 10$^\circ$C to 0$^\circ$C} & {Model with TL\tnote{1}} & {6.00} & {4.50} & {56.84} & {100} & {1min10sec} & {L(50)N(100)} & {$q=108, \alpha_1=0.25, \alpha_2=0.75$}\\
				& {Model without TL\tnote{2}} & {37.33} & {32.05} & {-} & {-} & {-} & {-} & {-} \\
				& {Accuracy improvement} & {82.00$\%$} & {85.96$\%$} & {-} & {-} & {-} & {-}  & {-} \\
				\cline{1-9}
				\multirow{3}*{From 10$^\circ$C to -10$^\circ$C} & {Model with TL\tnote{1}} & {15.80} & 12.41 & {24.79} & {100}  & {1min23sec} & {L(50)N(100)} & {$q=45, \alpha_1=0.11, \alpha_2=0.89$}\\
				& {Model without TL\tnote{2}} & {36.08} & {31.88} & {-} & {-} & {-} & {-}  & {-} \\
				& {Accuracy improvement} & {56.21$\%$} & {61.73$\%$} & {-} & {-} & {-} & {-}  & {-} \\
				\cline{1-9}
				\multirow{3}*{From 10$^\circ$C to -20$^\circ$C} & {Model with TL\tnote{1}} & {21.59} & {16.96} & {8.28} & {100} & {1min9sec} & {L(50)N(100)} & {$q=15, \alpha_1=0.035, \alpha_2=0.965$}\\
				& {Model without TL\tnote{2}} & {36.03} & {41.53} & {-} & {-} & {-} & {-}  & {-} \\
				& {Accuracy improvement} & {40.08$\%$} & {59.16$\%$} & {-}& {-} & {-} & {-}  & {-} \\
				\bottomrule
			\end{tabular}
			\begin{tablenotes}
				\footnotesize
				\item[1] denotes the developed reference SoC model at 10$^\circ$C using transfer learning at new temperatures;
				\item[2] denotes the developed reference SoC model at 10$^\circ$C directly applied at new temperatures;
				\item[-] denotes null.
			\end{tablenotes}
		\end{threeparttable}
	\end{center}
\end{table*}

\begin{figure}[!ht]
	\centering
	\includegraphics[scale=0.55]{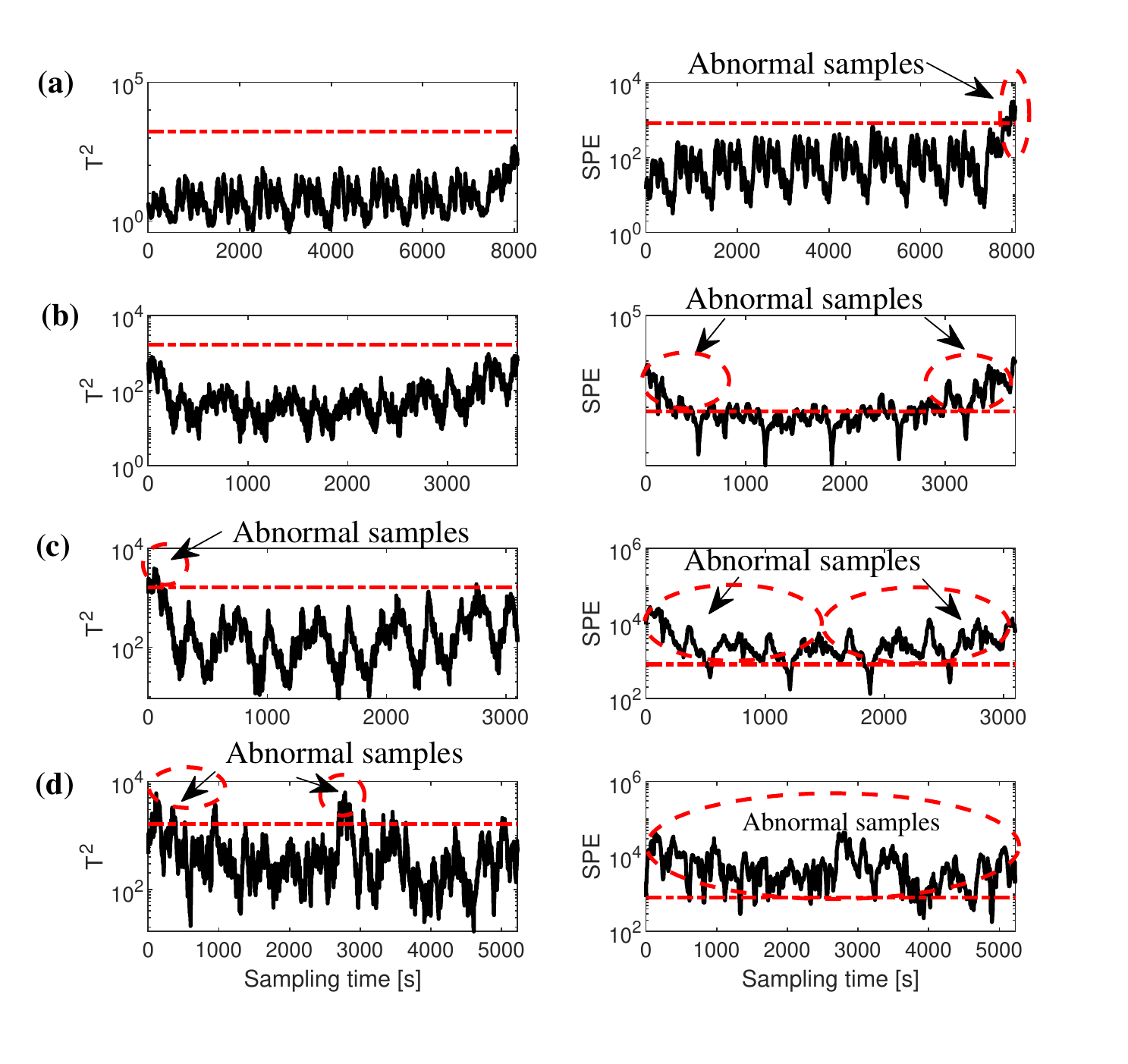}
	\caption{Monitoring results using US06 testing cycle at (a) 25$^\circ$C (b) 0$^\circ$C (c) -10$^\circ$C, and (d) -20$^\circ$C. (The red dashed line: control limits; the blue dotted line: monitoring statistics; the red dashed ellipse indicates abnormal samples)}
	\label{MyFig8}
\end{figure}

\begin{figure}[ht]
	\centering
	\includegraphics[scale=0.55]{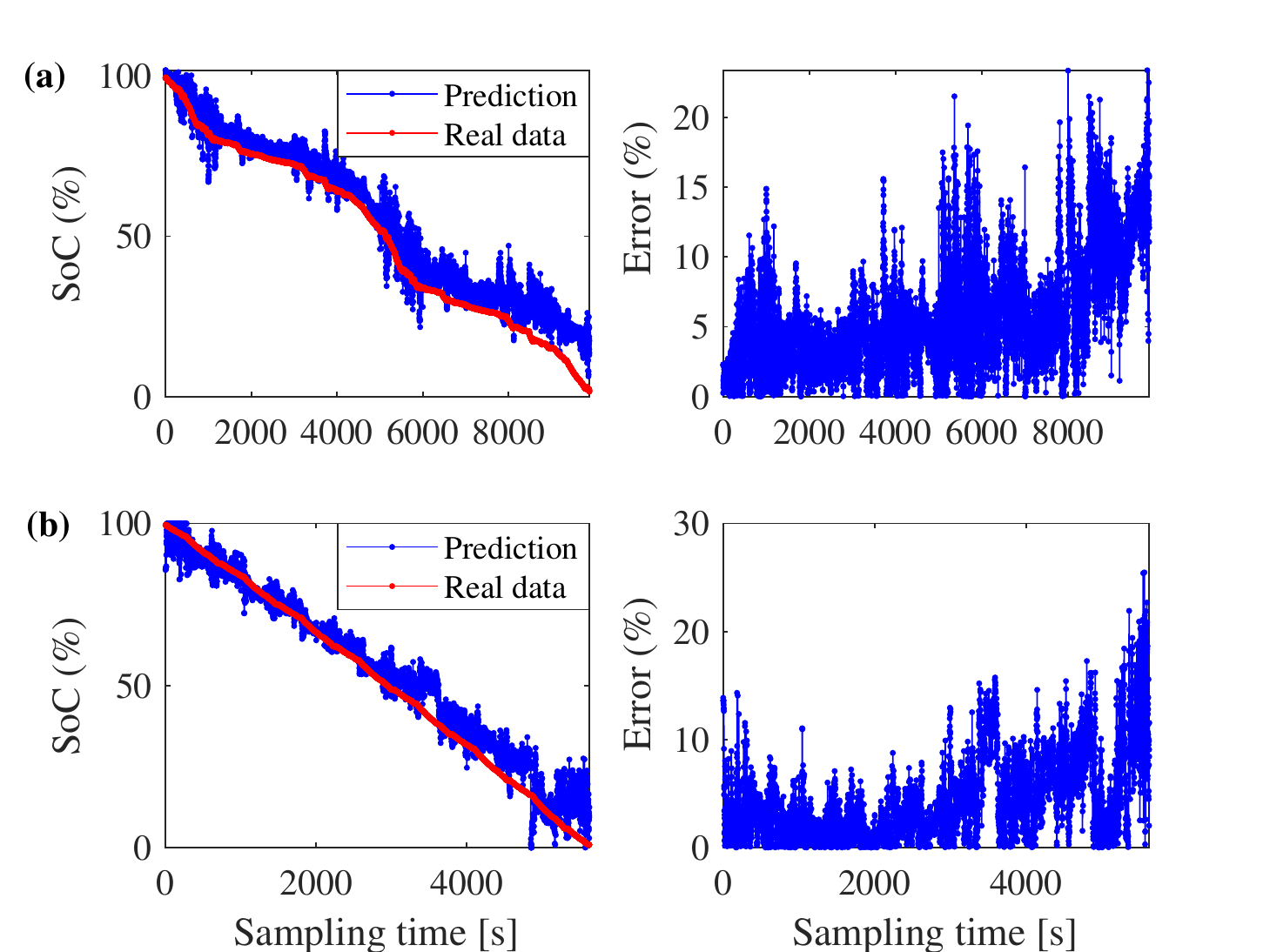}\\
	\caption{Estimation results using the proposed method with TL at (a) 25$^\circ$C and (b) 0$^\circ$C when the reference temperature is 10$^\circ$C.}
	\label{MyFig9}
\end{figure}

The network configurations are available in Table II. For each reference temperature, the initial learning rate in the Adam optimization algorithm is given as 0.01, and batch size is selected as 32. To explain the way for configuring networks, L(50,100)N(100) at the reference temperature 10$^\circ$C is taken as an example. It means that the network that has three hidden layers with 50 cells in the first LSTM layer, 100 cells in the second LSTM layer, and the third fully-connected layer with 100 nodes. Finally, there is an output layer with one node to output the SoC estimation. We use dropout and L2 regularization to control model overfitting. The employing of dropout in the fully-connected layer will prevent overfitting by ignoring randomly selected neurons during training, hence reducing the sensitivity to individual neurons' specific weights. A 20$\%$ dropout rate is recommended as a good compromise between retaining model accuracy and preventing overfitting. Besides, we stop in the training process earlier when there is no improvement in the validation data set. Besides the provided guidance for tuning the network, automatic hyperparameter tuning methods, such as Bayesian optimization [33], particle swarm optimization [34], are helpful to alleviate the tedious tuning procedure. For data at different reference temperatures, the network architecture would be slightly different.

\subsection{Predication Performance Evaluation}
In this part, the influence of temperature on prediction ability is investigated using the proposed method. Taking 10$^\circ$C as the reference temperature, a monitoring model is established according to the procedures given in Subsection III.B. Same as the last section, the first eight drive cycles are used for training and the following one cycle is used for validation to determine the crucial parameters, i.e., the number of retained CVs $R$. The parameter $R$ is determined as 163 as shown in Fig. 6(b), where a knee point is observed. Besides, the specific values of $R$ for the other temperatures to develop monitoring model can be found in Table II.

After the monitoring model is completed, the last discharge cycle at 10$^\circ$C is used as a testing case to validate the developed model, which is tested as normal but not shown here for brevity. Moreover, Fig. 8(a) to Fig. 8(d) visualize the monitoring results at the other temperatures. For 25$^\circ$C (Fig. 8(a)), it is observed that $T^2$ statistics remains below the control limit while $SPE$ statistic reaches an abnormal value at the end of the discharge cycle. For 0$^\circ$C (Fig. 8(b)), the $SPE$ monitoring statistics becomes obviously abnormal already at the beginning of the discharge cycle. For -10$^\circ$C (Fig. 8(c)) and -20$^\circ$C (Fig. 8(d)), more serious abnormalities are observed for both $T^2$ and $SPE$. Abnormal samples are circled by red dashed ellipses. Therefore, changes of temperature exert unignorable influences on estimation ability. The larger the difference between reference temperature and new temperatures is the more dissimilar temporal dynamics will be.

\begin{table*}[ht]
    \tiny
    \centering
	\renewcommand{\arraystretch}{1.2}
	\caption{Estimation results and similarities using the proposed method from the reference temperature -10$^\circ$C to temperature -20$^\circ$C.}
	\label{table_V}
	\begin{center}
		\begin{threeparttable}
			\begin{tabular}{c c c c| c c c| c}
				\toprule
				\multicolumn{1}{c}{\multirow{2}*{\textbf{Transferring case}}} & \multicolumn{3}{c|}{\textbf{Indices}} & \multicolumn{3}{c|}{\multirow{1}*{\textbf{Transfer network configuration}}} & \multicolumn{1}{c}{\multirow{2}*{\textbf{Specific values for parameters}}}\\
				\cline{2-7}
				& {$RMSE$} & {$MAE$} & {$H(\%$)} & {Iteration epochs} & {Training time} & {Network} \\
				\hline
				{From -10$^\circ$C to -20$^\circ$C} & {9.01} & {7.65} & {55.80} & {100} & {2min6sec} & {L(50)N(100)} & {$q=110, \alpha_1=0.25, \alpha_2=0.75$}\\
				{From 10$^\circ$C to -20$^\circ$C}  & {21.59} & {16.96} &	{8.28} & {100} & 1min9sec & {L(50)N(100)} & {$q=15, \alpha_1=0.035, \alpha_2=0.965$}\\
				{Accuracy improvement} & {58.27$\%$} & {54.89$\%$} & {-} & {-} & {-}  & {-} & {-}\\
				\bottomrule
			\end{tabular}
			\begin{tablenotes}
				\footnotesize
				\item[-] denotes null.
			\end{tablenotes}
		\end{threeparttable}
	\end{center}
\end{table*}

\subsection{Transfer SoC Estimation at Varying Temperatures}
Choosing 10$^\circ$C as the reference temperature, the transferability of the proposed method is verified for the other four temperatures concerning SoC estimation. Here, ten discharging cycling data with real SoC values at 10$^\circ$C and the first discharge cycle with real SoC values at the other temperatures are used as the training dataset for four transfer models, respectively. And the second discharge cycle at each temperature is used as corresponding testing data. After the data processing and feature extraction using CVA as shown in the last two sections, we will make progress to select the consistent temporal features between 10$^\circ$C and the other four temperatures. According to the steps given in Subsection III.C, the number of consistent temporal features at 25$^\circ$C, 0$^\circ$C, -10$^\circ$C, and -20$^\circ$C are 200, 108, 45, and 15, respectively. For each temperature, a shared prediction model is trained using consistent temporal features selected at the reference temperature, i.e., 10$^\circ$C. Thereafter, specific temporal features at new temperatures are separated from off-line discharge cycle and then specific prediction models are trained. Table III summarizes the estimation results of the proposed method with TL. Besides, estimation results by directly applying the reference model into the other temperatures are given for comparison, which as to be expected, lead to very large prediction errors. In contrast, using the proposed method with TL estimation errors have been greatly reduced by 79.48$\%$, 82.00$\%$, 56.21$\%$, and 40.08$\%$ at 25$^\circ$C, 0$^\circ$C, -10$^\circ$C, and -20$^\circ$C, respectively. Especially, for 25$^\circ$C and 0$^\circ$C, the estimation results are shown in Fig. 9 for an intuitive presentation. Although errors exist due to the limitation of data amount, the estimation trends meet well with the real case. Besides, more accurate results can be derived if filter algorithms are applied as post-processing. It should be noticed that the network configurations in Table III are tuned for specific prediction ability $F_t(\cdot)$ in Eq. (11).

In addition to the reduction of estimation error, the proposed method contributes to further process understanding. The similarity between reference temperature and new temperatures is quantitatively accounted by taking advantage of consistent CVs. The index $H={\operatorname{var}\left(\mathbf{Z}_{x,q}^{T} \mathbf{Z}_{x,q}\right)}/{\operatorname{var}\left(\mathbf{Z}_{x}^{T} \mathbf{Z}_{x}\right)}$ is defined to reveal the similarity from the perspective of process variances, where $\operatorname{var}(\cdot)$ calculates the variance of the matrix. A larger value of $H$ indicates a higher similarity. As shown in Table III, high similarities are observed for 25$^\circ$C and 0$^\circ$C, which are more than 50$\%$ with respect to $H$. As a result, a better transfer estimation ability is achieved at these two cases. In contrast, similarities are low for -10$^\circ$C and -20$^\circ$C, leading to relatively poor estimation results.

Next, the efficacy of the proposed method is further verified through a transfer procedure from -10$^\circ$C to -20$^\circ$C. Choosing -10$^\circ$C as the reference temperature, the number of consistent temporal correlations between -20$^\circ$C and -10$^\circ$C is 110 using the first discharge cycle at -20$^\circ$C for training. The similarities between these two temperatures are calculated as shown in Table \ref{table_V}, which is 55.80$\%$. For better comparison, the results obtained from 10$^\circ$C to -20$^\circ$C are summarized in Table IV. Using the second discharge cycle at -20$^\circ$C as the testing data, a higher similarity is observed at the new reference temperature, i.e., -10$^\circ$C, and this contributes to estimation error reduced by as much as 58.27$\%$ for $RMSE$ in comparison with the result when the reference temperature is 10$^\circ$C. Similar results can be obtained with respect to index $MAE$.

\section{Conclusion}
In this article, a data-driven SoC estimation framework of Lithium-ion batteries is designed to consider the influences of varying ambient temperatures. The accuracy of reference SoC estimation model based on long short-term memory network has been greatly improved by taking advantage of temporal dynamics. Furthermore, it is more practical and effective in comparison with other algorithms through two achievements: 1) evaluating the influences of temperatures on SoC estimation ability by checking the variability of temporal features; 2) rapid updating the reference SoC estimation using advanced transfer learning when the estimation relationship has been changed. In this way, quantitative indicators contribute to automatically indicate the efficacy of the reference SoC estimation model. With a small number of cycles at new temperatures, estimation results are achieved by inheriting the consistent estimation ability. The proposed approach is exposed to extensive testing provided by a public benchmark and it has proven to be a powerful tool for SoC estimation without the limitation of battery type and material.

Exploiting transfer learning ability, the proposed method enables accurate SoC estimation from the reference temperature to new temperatures. However, efficacy of the proposed method may be challenged if the temperature experiences a large-scale variation, resulting in that new temperatures are far away from all reference temperatures. It may be promising to overcome the aforementioned challenge by introducing a hybrid estimation model with accessible domain knowledge.

\bibliographystyle{Bibliography/IEEEtranTIE}

\begin{thebibliography}{25}
	\bibitem{Ref1}
	N. Tian, H. Fang, and Y. Wang, ``Real-time optimal lithium-ion battery charging based on explicit model predictive control,"  \textit{IEEE Trans. Ind. Informat.}, vol. 17, no. 2, pp. 1318-1330, Feb. 2021.
	
	\bibitem{Ref2}
	X. Hu, C. Zou, C. Zhang, and Y. Li, ``Technological developments in batteries: a survey of principal roles, types, and management needs,'' \textit{IEEE Power Energy Mag.}, vol. 15, no. 5, pp. 20-31, 2017.
	
	\bibitem{Ref3}
	Y.J. Xing, W. He, M. Pecht, and K.L. Tsui, ``State of charge estimation of lithium-ion batteries using the open-circuit voltage at various ambient temperatures,'' \textit{Appl. Energy}, vol. 113, pp. 106-115, 2014.
	
	\bibitem{Ref4}
	S.M. George, I.D. Dimitrios, T.A. Papadopoulos, D.P. Labridis, and V.G. Agelidis, ``State-of-charge estimation for Li-ion batteries: a more accurate hybrid approach,'' \textit{IEEE Trans. Energy Convers.}, vol. 34, no. 1, pp. 109-119, 2019.
	
	\bibitem{Ref5}
	F. Zhang, G. Liu, L. Fang, and H. Wang, ``Estimation of battery state of charge with H1 observer: Applied to a robot for inspecting power transmission lines,'' \textit{IEEE Trans. Ind. Electron.}, vol. 59, no. 2, pp. 1086-1095, 2012.
	
	\bibitem{Ref6}
	W. He, N. Williard, C. Chen, and M. Pecht, ``State of charge estimation for Li-ion batteries using neural network modeling and unscented Kalman filter-based error cancellation,'' \textit{Int. J. Elec. Power}, vol. 62, pp. 783-791, 2014.
	
	\bibitem{Ref7}
	C.R. Li, F. Xiao, and Y.X. Fan, ``An approach to state of charge estimation of Lithium-ion batteries based on recurrent neural networks with gated recurrent unit,'' \textit{Energies}, vol. 12, pp. 1592-1603, 2019.
	
	\bibitem{Ref8}
	E. Chemali, P.J. Kollmeyer, M. Preindl, R. Ahmed, and A. Emadi, ``Long short-term memory networks for accurate state-of-charge estimation of Li-ion batteries,'' \textit{IEEE Trans. Ind. Electron.}, vol. 65, no. 8, pp. 6730-6739, 2018.
	
	\bibitem{Ref9}
	Y. Wang, C. Zhang, and Z. Chen, ``A method for state-of-charge estimation of LiFePO4 batteries at dynamic currents and temperatures using particle filter,'' \textit{J. Power Sources}, vol. 279, pp. 306-311, 2015.
	
   	\bibitem{Ref10}
    	S. Ma, M. Jng, P. Tao, C.Y. Song, J.B. Wu, J. Wang, T. Deng, and W. Shang, ''Temperature effect and thermal impact in lithium-ion batteries: A review,'' \textit{Prog. Nat. Sci.}, vol. 28, no. 6, pp. 653-666, 2018.
    	
	\bibitem{Ref11}
	D. Huang, Z. Chen, C. Zheng, and H. Li, ``A model-based state-of-charge estimation method for series-connected lithium-ion battery pack considering fast-varying cell temperature,'' \textit{Energy}, vol. 185, pp. 847-861, 2019.

	\bibitem{Ref12}
	X. Tang, Y. Wang, C. Zou, K. Yao, Y. Xia, and F. Gao, ``A novel framework for Lithium-ion battery modeling considering uncertainties of temperature and aging,'' \textit{Energy Convers. Manag.}, vol. 180, pp. 162-170, 2019.
	
	\bibitem{Ref13}
	S.J. Pan, and Q. Yang, ``A survey on transfer learning,'' \textit{IEEE Trans. Knowl. Data En.}, vol. 22, no. 10, pp. 1345-1359, 2010.
	
	\bibitem{Ref14}
	P. Kollmeyer, ``Panasonic 18650PF Li-ion battery data,'' Mendeley data. v1. 2018. http://dx.doi.org/10.17632/wykht8y7tg.1.
	
	\bibitem{Ref15}
	P. Wu, S. Lou, X. Zhang, J. He, Y. Liu, and J. Gao, ``Data-Driven fault diagnosis using deep canonical variate analysis and Fisher discriminant analysis," \textit{IEEE Trans. Ind. Informat.}, doi: 10.1109/TII.2020.3030179.
	
	\bibitem{Ref16}
	B.B. Jiang, X.X. Zhu, D.X. Huang, and R.D. Braatz, ``Canonical variate analysis-based monitoring of process correlation structure using causal feature representation,'' \textit{J. Process Contr.}, vol. 32, pp. 109-116, 2015.
	
	\bibitem{Ref17}
	Y.Z. Zhang, R. Xiong, H.W. He, and M.G. Pecht, ``Long short-term memory recurrent neural network for remaining useful life prediction of Lithium-ion batteries,'' \textit{IEEE Trans. Veh. Technol.}, vol. 67, no. 7, pp. 5695-5705, 2018.
	
    \bibitem{Ref18}
	A. Grossmann, and J. Morlet, ``Decomposition of Hardy functions into square integrable wavelets of constant shape,'' \textit{SIAM J. Math. Anal.}, vol. 15, no. 4, pp. 723-736, 1984.

	\bibitem{Ref19}
	D. Liu, J. Zhou, H. Liao, Y. Peng, and X. Peng, ``A health indicator extraction and optimization framework for lithium-ion battery degradation modeling and prognostics," \textit{ IEEE Trans. Syst., Man, Cybern. Syst.}, vol. 45, no. 6, pp. 915-928, 2015.

	\bibitem{Ref20}
	Y.P. Zhou, M.H. Huang, Y.P. Chen, and Y. Tao, ``A novel health indicator for on-line lithium-ion batteries remaining useful life prediction," \textit{J. Power Sources}, vol. 321, pp. 1-10, 2016.

	\bibitem{Ref21}
	Y.G. Lei, N.P. Li, L. Guo, N.B. Li, T. Yan, and J. Lin, ``Machinery health prognostics: A systematic review from data acquisition to RUL prediction," \textit{Mech. Syst. Signal Process}, vol. 104, pp. 799-834, 2018.

	\bibitem{Ref32}
  	T. Guo, T. Lin, and N. Antulov-Fantulin, ``Exploring interpretable LSTM neural networks over nulti-variable data," in \textit{Proc. 36th Int. Conf. Mach. Learn.}, vol. 97, California USA, 9-15 Jun., 2019, pp. 2494-2504.

	\bibitem{Ref22}
	D.P. Kingma, and L.J. Ba, ``Adam: A method for stochastic optimization event,'' in \textit{Proc. Int. Conf. Learn. Represent.}, San Diego USA,  7-9 May, 2015, pp. 1-13.

	\bibitem{Ref23}
	Y. Qin, and C.H. Zhao, ``Comprehensive process decomposition for closed-loop process monitoring with quality-relevant slow feature analysis,'' \textit{J. Process Contr.}, vol. 77, pp. 141-154, 2019.

    	\bibitem{Ref24}
	B. Song, H. Shi, S. Tan, and Y. Tao, ``Multi-subspace orthogonal canonical correlation analysis for quality related plant wide process monitoring,''  \textit{IEEE Trans. Ind. Informat.}, doi: 10.1109/TII.2020.3015034.
 	
	\bibitem{Ref25}
	 S.S. Shapiro, and M.B. Wilk, ``An analysis of variance test for normality (complete samples)," \textit{Biometrika}, vol. 52, no. (3-4), pp. 591-611, 1965.

	\bibitem{Ref26}
	C. Shang, F. Yang, B. Huang, and D. Huang, ``Recursive slow feature analysis for adaptive monitoring of industrial processes," \textit{IEEE Trans. Ind. Electron.}, vol. 65, no. 11, pp. 8895-8905, 2018.

	\bibitem{Ref27}
           K.E.S. Pilario, Y. Cao, and M. Shafiee, ``Mixed kernel canonical variate dissimilarity analysis for incipient fault monitoring in nonlinear dynamic processes," \textit{Comput. Chem. Eng.}, vol. 123, no. 6, pp. 143-154, 2019.
		
	\bibitem{Ref28}
	P.L. Zhao, C.H.H. Steven, J.L. Wang, and L.I. Bin, ``Online transfer learning,'' \textit{Artif. Intell.}, vol. 216, pp. 76-102, 2014.
	
	\bibitem{Ref29}
	R. Razavi-Far, S. Chakrabarti, M. Saif, and E. Zio, ``An integrated imputation-prediction scheme for prognostics of battery data with missing observations,"   \textit{Expert Syst. Appl.}, vol. 115, pp. 709-723, 2019.

	\bibitem{Ref30}
	R.R. Richardson, C.R. Birkl, M.A. Osborne, and D.A. Howey, ``Gaussian process regression for in situ capacity estimation of lithium-ion batteries," \textit{IEEE Trans. Ind. Informat.}, vol. 15, no. 1, pp. 127-138, 2019.

	\bibitem{Ref31}
	 K.E.S. Pilario, and Y. Cao, ``Canonical variate dissimilarity analysis for process incipient fault detection,"  \textit{IEEE Trans. Ind. Informat.}, vol. 14, no. 12, pp. 5308-5315, 2018.

	\bibitem{Ref33}
	J. Snoek, H. Larochelle, and R. Adams, ``Practical Bayesian optimization of machine learning algorithms,'' in \textit{Adv. Neural Inf. Process. Syst.}, pp. 2960-2968, 2012.

	\bibitem{Ref34}
	D. Liu, L. Li, Y. Song, L. Wu, and Y. Peng, ``Hybrid state of charge estimation for lithium-ion battery under dynamic operating conditions,'' \textit{Int. J. Electr. Power Energy Syst.}, vol. 110, pp. 48-61, 2019.
\end{thebibliography}

\end{document}